\documentclass[12pt]{article}

\usepackage{arxiv}

\usepackage[utf8]{inputenc} 
\usepackage[T1]{fontenc}    
\usepackage{hyperref}       
\usepackage{url}            
\usepackage{booktabs}       
\usepackage{amsfonts}       
\usepackage{nicefrac}       
\usepackage{microtype}      
\usepackage{lipsum}
\usepackage{graphicx}
\usepackage{amsmath,amsfonts,amssymb}
\usepackage{graphicx}
\usepackage{subcaption}
\usepackage{color}

\title{Higher order quantum reservoir computing for non-intrusive reduced-order models}

\author{
 Vinamr Jain \\
  Department of Physics, \\
  Indian Institute of Technology, Delhi\\
  Hauz Khas, New Delhi, Delhi-110016, India \\
  \texttt{ph1200734@iitd.ac.in}
   \And
  Romit Maulik \\
  College of Information Sciences and Technology\\
  Pennsylvania State University\\
  University Park, PA 16802, USA. \\
  \texttt{rmaulik@psu.edu} \\
}

\begin{document}
\maketitle
\begin{abstract}
Forecasting dynamical systems is of importance to numerous real-world applications. When possible, dynamical systems forecasts are constructed based on first-principles-based models such as through the use of differential equations. When these equations are unknown, non-intrusive techniques must be utilized to build predictive models from data alone. Machine learning (ML) methods have recently been used for such tasks.  Moreover, ML methods provide the added advantage of significant reductions in time-to-solution for predictions in contrast with first-principle based models. However, many state-of-the-art ML-based methods for forecasting rely on neural networks, which may be expensive to train and necessitate requirements for large amounts of memory. In this work, we propose a quantum mechanics inspired ML modeling strategy for learning nonlinear dynamical systems that provides data-driven forecasts for complex dynamical systems with reduced training time and memory costs. This approach, denoted the quantum reservoir computing technique (QRC), is a hybrid quantum-classical framework employing an ensemble of interconnected small quantum systems via classical linear feedback connections. By mapping the dynamical state to a suitable quantum representation amenable to unitary operations, QRC is able to predict complex nonlinear dynamical systems in a stable and accurate manner. We demonstrate the efficacy of this framework through benchmark forecasts of the NOAA Optimal Interpolation Sea Surface Temperature dataset and compare the performance of QRC to other ML methods.
\end{abstract}


\section{Introduction}
Quantum computing has emerged as a promising frontier offering exponential speed-up potential for solving complex problems compared to classical computing methods. The distinctive ability of quantum computers to efficiently process large datasets, simulate quantum systems, and optimize solutions has spurred significant interest and research. Variational quantum algorithms have shown promise to be utilized for computational fluid dynamics applications \cite{jaksch2023variational}. Quantum machine learning (QML) has particularly garnered attention as an interdisciplinary research area, leveraging quantum mechanics' inherent properties to potentially surpass classical machine learning methods. 
The integration of quantum computing (QC) and machine learning (ML) in dynamical systems offers promising advancements in forecasting capabilities. Quantum algorithms can handle high-dimensional data more efficiently, improving the accuracy of dynamical systems models. Concurrently, ML techniques enhance these models' interpretability and predictive power by leveraging large datasets and complex patterns \cite{givi2021machine}. 
Quantum algorithms have demonstrated polynomial speedups for both linear and nonlinear differential equations. These speedups become exponentially larger as the problem dimensions increase, showing promise for applications in computational fluid dynamics (CFD) and other areas \cite{givi2020quantum}. 
Adiabatic annealing-based quantum computing methods have been studied for solving fluid dynamics problems such as the Navier-Stokes (NS) equation \cite{ray2019towards}. However, Lewis et al. (2023) \cite{lewis2023limitations} suggests that simulating the classical dynamics of turbulent or chaotic systems may not offer an exponential quantum advantage over current classical algorithms. 
Succi et al. (2024) \cite{succi2024ensemble} investigated the feasibility of using the Liouville formulation for ensemble simulations of fluid flows on quantum computers. However, the Liouville approach is anticipated to necessitate several hundred logical qubits, which is limited by current hardware. 
The inherently non-local nature of quantum mechanics might be beneficial for accurately representing the classical non-locality observed in low-Reynolds number flows \cite{succi2023quantum}. 
Sanavio et al. (2024) \cite{sanavio2024lattice} propose a quantum computing algorithm for fluid flows based on the Carleman-linearization
of the Lattice Boltzmann (LB) method.

Freeman et al. (2024) \cite{freeman2024quantum} introduced a novel data-driven parameterization scheme for unresolved components of dynamical systems using quantum mechanics and Koopman operator theory, successfully applied to the Lorenz 63 and 96 systems.
However, the method's scalability and practicality in complex, higher-dimensional systems remain to be fully assessed. Giannakis et al. (2022) \cite{giannakis2022embedding} developed a framework for simulating measure-preserving, ergodic dynamical systems on quantum computers by combining ergodic theory with quantum information science. The framework shows promise for developing data-driven quantum algorithms for modelling nonlinear dynamics and forecasting real-world systems, but further exploration is needed for applications such as climate dynamics and turbulent fluid flows.
Gourianov et al. (2022) \cite{gourianov2022quantum} utilize interscale correlations in turbulent flows to develop a structure-resolving simulation algorithm based on Matrix Product States (MPS). This quantum-inspired method significantly reduces computational complexity and shows potential for future implementation on quantum computers. 

Pfeffer et al. (2022) explored the use of a hybrid quantum-classical gate-based reservoir computing model to predict and reconstruct the dynamics of classical, nonlinear thermal convection flows \cite{pfeffer2022hybrid}. The study demonstrated that the quantum reservoir, utilizing entangled qubits, can match the performance of classical reservoirs that use thousands of perceptrons, even when implemented on a noisy intermediate-scale quantum (NISQ) device. However, it's important to note that the model was tested on relatively low-dimensional systems. Extending this approach to higher-dimensional and more complex systems, such as turbulent flows, and real-world applications, remains a challenge.

Pfeffer et al. (2023) \cite{pfeffer2023reduced} proposed hybrid quantum-classical gate-based reservoir computing models to replicate low-order statistical properties of a two-dimensional turbulent Rayleigh-Benard convection flow. These models aim to learn the nonlinear and chaotic dynamics of the turbulent flow in a lower-dimensional latent data space, using time-dependent expansion coefficients of the most energetic Proper Orthogonal Decomposition (POD) modes. The training data is generated by conducting a POD-based compression of the direct numerical simulations of the original turbulent flow. The hybrid quantum reservoir computing models operate in a reconstruction mode, taking 3 POD modes as input at each step and reconstructing the missing 13 modes. The quantum models perform similarly to the classical model with the added cost of sophisticated quantum hardware. We are inspired by this study to develop a non-intrusive reduced-order modeling strategy that may offer significant benefits when the underlying governing equations are unavailable - and therefore not amenable to classical physics-based modeling.


Machine learning (ML) methods, and neural networks in particular, are an attractive approach for the computationally efficient forecasting of various realistic time-series data offering a data-driven alternative to traditional numerical methods which struggle with rapid prediction and real-time control in applications like weather and climate among others. Typical ML baselines include the long short-term memory network (LSTMs) \cite{hochreiter1997long} and Gated Recurrent Units (GRUs) \cite{chung2014empirical}. They are both types of recurrent neural networks designed for sequence data without the problem of vanishing gradients, but they differ in their architectures and how they handle information flow. LSTMs are equipped with a more complex architecture that includes three gates: input, output, and forget gates, along with a memory cell that allows for better long-term memory retention and control over information flow. GRUs simplify this structure by using only two gates: reset and update gates, merging the cell state and hidden state into one. While LSTMs offer more control and flexibility due to their more intricate gating mechanisms, GRUs are computationally more efficient and easier to implement, often providing comparable performance in practice.

In this study, we investigate the use of data-driven methods to perform forecasts for a geophysical dataset given by the sea-surface temperature (SST). SST prediction is a commonly chosen benchmark for assessing the quality of a data-driven forecasting model. For example, Yang et al. (2017) \cite{yang2017cfcc} use CFCC-LSTMs and a 3D grid to deal with spatiotemporal information to predict future SST values. Zhang, Qin et al. (2017) \cite{zhang2017prediction} formulated the prediction of SST as a time series regression problem and proposed an LSTM-based network to model the temporal relationship of SST in order to predict future values.
Zhang, Zhen (2020)\cite{zhang2020monthly} proposed a fully connected neural network model based on GRU is designed to predict SST over the medium and long term.
 Xie et al. (2019)\cite{xie2019adaptive} proposed a gate recurrent unit (GRU) encoder-decoder with
SST codes and dynamic influence links (DIL) to predict SST for long-term predictions.

Reservoir Computing (RC) is a class of data-driven forecasting frameworks in which a fixed, randomly connected network, known as a reservoir, processes input streams nonlinearly. The processed data is then fed into a trainable readout for pattern analysis and processing of temporal or sequential data. Reservoir Computing (RC) has demonstrated significant success in modeling the dynamics of high-dimensional chaotic systems \cite{vlachas2020backpropagation}.

Quantum Reservoir Computing (QRC) extends this paradigm by implementing the reservoir using quantum many-body systems, leveraging quantum dynamics for information processing. This adaptation opens avenues for exploring quantum advantages in temporal machine-learning tasks. In this context, Higher Order Quantum Reservoir Computing (HQRC) introduces a novel approach by organizing quantum systems into an ensemble reservoir. Each system within this ensemble, akin to a computational node, receives inputs and signals from other systems, enhancing computational power through controllable linear feedback connections. This higher-order scheme offers increased expressiveness and scalability, potentially unlocking advanced capabilities for temporal forecasting tasks.

This paper focuses on demonstrating the efficacy of Higher Order Quantum Reservoir Computing (HQRC) framework proposed by Tran and Nakajima (2020)\cite{tran2020higherorder} in forecasting Sea Surface Temperature (SST) data. By leveraging the unique features of quantum systems within an ensemble reservoir framework, we aim to showcase the potential of HQRC for addressing real-world prediction challenges. Through empirical evaluation and comparative analysis, we highlight the advantages of HQRC for SST forecasting, contributing to the growing body of research on quantum-enhanced machine learning techniques.

\section{Methods and data}

\subsection{Quantum Reservoir Dynamics}

For ease of exposition and without loss of generality, let us consider a one-dimensional input and output case with the input sequence of scalars \(u = \{u_1, \ldots, u_L\}\) and the corresponding target sequence \(\hat{y} = \{\hat{y}_1, \ldots, \hat{y}_L\}\), where \(u_k\) is a continuous variable in \([0, 1]\). Quantum Reservoir Computing (QRC) emulates a nonlinear function \(Q\) to produce the output \(y_k = Q(w, \rho^{(0)}, \{u_l\}_{l=1}^k)\) (\(k = 1, \ldots, L\)). Here, \(\rho^{(0)}\) is the initial state of the quantum system, and \(w\) is the parameter that needs to be optimized.

A temporal learning task consists of three phases: a washout phase, a training phase, and an evaluation phase. In the washout phase, the system evolves for the first \(T\) transient steps to wash out the initial conditions from the dynamics. The training phase to optimize \(w\) is performed with training data \(\{u_k\}_{k=T}^{L_1}, \{y_k\}_{k=T}^{L_1}\), where \(1 \leq T < L_1 < L\), such that the mean-square error between \(y_k\) and \(\hat{y}_k\) over \(k = T, \ldots, L_1\) becomes minimum. The trained parameter \(w\) is used to generate outputs in the evaluation phase.

For an \(N\)-qubits system, at time \(t = (k-1)\tau\), the input \(u_k \in [0, 1]\) is fed to the system by setting the density matrix of the first spin to
\begin{align}
\rho_{u_k} = (1 - u_k)|0\rangle\langle 0| + u_k |1\rangle\langle 1| \in \mathbb{M}_{2 \times 2}.    
\end{align}
Therefore, the density matrix \(\rho \in \mathbb{M}_{2^N \times 2^N}\) of the entire system is mapped by a completely positive and trace preserving (CPTP) map
\begin{align}
    \rho \to \mathcal{T}_{u_k}(\rho) = \rho_{u_k} \otimes \text{Tr}_1[\rho],     
\end{align}
where \(\text{Tr}_1\) denotes a partial trace with respect to the first qubit. After the input is set, the system continues evolving itself during the time interval \(\tau\). The dynamics are governed by the Schrödinger equation, and the information of the input sequence encoded in the first spin spreads through the system. It follows that the state of the system before the next input \(u_{k+1}\) is
\begin{align}
\rho^{(k)} = e^{-iH\tau}\mathcal{T}_{u_k}(\rho^{(k-1)})e^{iH\tau},    
\end{align}

where \(\rho^{(k)} = \rho(k\tau)\) is the density matrix at \(t = k\tau\).

If we employ the ordered basis \(\{O_j\}\) in the operator space, then the observed signals at time \(t\) are the first \(N_{\text{out}}\) elements
\begin{align}
s_j(t) = \text{Tr}[\rho(t)O_j],
\end{align}
where the selection of observables depends on the physical implementation of the system. In quantum mechanics, the ordered basis \(\{O_j\}\) in the operator space typically consists of a set of linearly independent operators that span the space of all observables for a given system. For example, in the case of a single qubit, a common ordered basis consists of the Pauli matrices \(\{I, \sigma_x, \sigma_y, \sigma_z\}\), where \(I\) is the identity matrix, and \(\sigma_x, \sigma_y, \sigma_z\) are the Pauli matrices. These matrices form a complete basis for the space of \(2 \times 2\) Hermitian operators, meaning any observable for a single qubit can be expressed as a linear combination of these basis operators. The choice of basis depends on the physical implementation and measurement capabilities of the system.

\subsection{Temporal Multiplexing}
A temporal multiplexing scheme is introduced to enhance performance in extracting dynamics. In this scheme, signals are measured not only at time \(k\tau\) but also at each of the subdivided \(V\) time intervals during the evolution in the interval \(\tau\) to construct \(V\) virtual nodes. The density matrix is then updated by
\begin{align}
\rho((k - 1)\tau + \frac{1}{V}\tau) = U_{\left(\frac{\tau}{V}\right)} T_{u_k}(\rho^{(k-1)}) U^\dagger_{\left(\frac{\tau}{V}\right)},
\end{align}
\begin{align}
\rho\left((k - 1)\tau + \frac{v}{V}\tau\right) = U_{\left(\frac{\tau}{V}\right)} \rho\left((k - 1)\tau + \frac{v - 1}{V}\tau\right) U^\dagger_{\left(\frac{\tau}{V}\right)}, \quad (v = 2, \ldots, V),
\end{align}
where \(U_{(\tau/V)} = e^{-iH(\tau/V)}\). Therefore, temporal signals from \(N_{\text{out}}V\) nodes can be obtained.
The learning procedure is straightforward, parameterizing the linear readout function as \(y_k = \sum_{i=0}^{N_{\text{out}}V} w_i x_{ki}\), where \(x_{ki} = s_j((k - 1)\tau + \frac{v}{V}\tau)\) for \(i = (j - 1)V + v > 0\) (\(1 \leq v \leq V, 1 \leq j \leq N_{\text{out}}\)). Here, \(x_{k0} = 1.0\) are introduced as constant bias terms, and \(w = [w_0, w_1, \ldots, w_{N_{\text{out}}V}]\) represents the readout weight parameters. If we denote \(K\) as the number of time steps used in the training phase, \(w\) is optimized via linear regression, or Ridge regression in the matrix form
\begin{align}
\hat{w}^T = (X^T X + \beta I)^{-1} X^T \hat{y},
\end{align}
where \(\hat{y} = [y_1, \ldots, y_K]^T\) is the target sequence, \(X = [x_{ki}] \in \mathbb{R}^{K \times (N_{\text{out}}V + 1)}\) is the training data matrix, and \(\beta\) is the parameter serving as the positive constant shifting the diagonals introduced to avoid the problem of the near-singular moment matrix.


\subsection{Higher Order Quantum Reservoir Computing model}
In the paper, The fully connected transverse field Ising model, a standard framework for building Quantum Reservoirs (QR), is employed. The Hamiltonian is given by \(H = J \sum_{i \neq j} h_{i,j}\sigma^x_{i}\sigma^x_{j} + J \sum_{j} g_j\sigma^z_{j}\), where \(\sigma^\gamma_{ j}\) (\(\gamma \in \{x, y, z\}\)) is the operator measuring the spin \(j\) along the \(\gamma\) direction. This operator can be described as an \(N\)-tensor product of \(2 \times 2\)-matrices:
\begin{align}
\sigma^\gamma_{j} = I \otimes \ldots \otimes \sigma^\gamma_{j} \otimes \ldots \otimes I, 
\end{align}
where \(I = \begin{bmatrix} 1 & 0 \\ 0 & 1 \end{bmatrix}\), \(\sigma_x = \begin{bmatrix} 0 & 1 \\ 1 & 0 \end{bmatrix}\), \(\sigma_y = \begin{bmatrix} 0 & -i \\ i & 0 \end{bmatrix}\), and \(\sigma_z = \begin{bmatrix} 1 & 0 \\ 0 & -1 \end{bmatrix}\). \(J\) is the coupling magnitude of the Hamiltonian, while the coupling parameter \(h_{i,j}\) and the transverse field parameter \(g_j\) are uniformly selected from the range \([-1.0, 1.0]\). \(N_{\text{out}} = N\) observables \(O_j = \sigma^z_{j}\) are chosen to produce the signals of the readout nodes. 
A Higher-Order Quantum Reservoir (HQR) comprises an ensemble of \(N_{\text{qr}}\) Quantum Reservoirs (QRs). The \(l\)-th system \(Q_l\) has the Hamiltonian \(H_l\) with \(N_l\) qubits (known as true nodes) and \(V_l\) virtual nodes. In order to interface between an input of dimension \(N_{\text{in}}\), we tile the input to match the HQRC dimension (Hence, ensuring \(N_{\text{in}}\) divides $N_{\text{qr}}$).
To simplify, we set all parameters the same for each system, i.e. \(H_l = H\), \(N_l = N\) and \(V_l = V\). Each individual reservoir is then evolved independently under a temporal multiplexing scheme as described above and the final readouts at each time step \(t = k\tau\)
are compiled as a VN dimensional vector \(x_{k} = [x_{k1}, x_{k2} \ldots, x_{kVN}]\). 

The reservoir states of \(Q_l\) at time \(t = k\tau\) (\(k \geq 0\)) are thus each represented by a \(V_lN_l\)-dimensional vector \(x_{kl}\), initialized at \(k = 0\) as the zero vector \(x_0\). Then, the reservoir states of the HQR at \(t = k\tau\) can be represented by an \(N_{\text{total}}\)-dimensional vector \(z_k = [x_{k1}^T, \ldots, x_{kN_{\text{qr}}}^T]^T\), where \(N_{\text{total}} = \sum_{l=1}^{N_{\text{qr}}} N_lV_l\). The scaling \(z \to (z + 1)/2\) is employed, such that the elements of \(z\) are in \([0, 1]\).

At \(t = (k - 1)\tau\) (\(k \geq 1\)), the mixed input \(u'_{kl}\), injected into \(Q_l\), is a linear combination of the external input and the reservoir states from all QRs. The mixed input \(u'_{k} \in \mathbb{R}^{N_{\text{qr}} \times 1}\) can be represented in matrix form as
\begin{align}
u'_{k} = (1 - \alpha)W_{\text{in}}u_k + \alpha W_{\text{con}}z_{k-1}, 
\end{align}
where \(W_{\text{con}} \in \mathbb{R}^{N_{\text{qr}} \times N_{\text{total}}}\) is randomly generated and fixed to represent the linear feedback connection between the QRs, and \(\alpha\) (\(0 \leq \alpha \leq 1\)) is defined as the connection strength parameter. Note that, each reservoir is fed one input dimension and the final construction of $z_{k}$ represents information from all the dimensions.

After injecting the mixed input \(u'_{kl}\) into \(Q_l\), the density matrix in \(Q_l\) is transformed by the CPTP map, \(T_{u'_{kl}}\), and consequently evolves in each \(\tau/V\) time.

For the state evolution of the system to be replicated without requiring feeding the output from the readout part into each QR’s input, Let $W_{\text{out}} \in \mathbb{R}^{N_{\text{out}} \times N_{\text{total}}}$ denote the trained readout weights. In the case of forecasting, $N_{\text{out}} = N_{\text{in}}$. We replicate $W_{\text{out}}$ into $W'_{\text{out}} \in \mathbb{R}^{N_{\text{qr}} \times N_{\text{total}}}$ (this is possible since $N_{\text{qr}}$ is a multiple of $N_{\text{in}}$) and represent the input in the closed system when the external input $u_k$ is withdrawn as
\begin{align}
u'_{k} = (1 - \alpha)W'_{\text{out}}z_{k-1} + \alpha W_{\text{con}}z_{k-1},
\end{align}
where $z_{k-1} \in \mathbb{R}^{N_{\text{total}} \times 1}$ is the reservoir states of HQR at $t = (k - 1)\tau$. The predicted output is given by 
\begin{align}
u^{\text{pred}}_{k} = W'_{\text{out}}z_{k-1}
\end{align}
We need to ensure that $u^{\text{pred}}_{k}$ does not fall out of the range [0,1] and so we perform clipping to restrict the prediction within this range.
\begin{align}
u_k = 
\begin{cases} 
1 & \text{if } u_k > 1 \\ 
u_k & \text{if } 0 \leq u_k \leq 1 \\ 
0 & \text{if } u_k < 0 
\end{cases}
\end{align}
The number of parameters is given by the size of the Readout matrix, shape $(N_{\text{total}}+1)\times N_{\text{qr}}$, where $N_{\text{total}}=\sum_{l=1}^{N_{\text{qrc}}}N_lV_l$. For generalizing to higher dimensional cases, the aforementioned procedure is performed for each scalar component of the state-vector and $W_{in}$ is learned jointly to account for interactions across components. Note that $W_{con}$ is not learnt and is fixed randomly.


\subsection{The sea-surface temperature dataset}

In this study, we use an open-source repository for our training and test data. Specifically, we utilize the National Oceanic and Atmospheric Administration (NOAA) Optimum Interpolation sea surface temperature V2 data set (henceforth NOAA-SST).\footnote{https://www.esrl.noaa.gov/psd/}  This data set exhibits a periodic behavior due to seasonal cycles in addition to fine-scaled phenomena emerging from complex ocean dynamics. Weekly-averaged NOAA-SST data are available on a quarter-degree grid which is sub-sampled to a one-degree grid for the purpose of our demonstrations. This data set has previously been used in several data-driven studies (for instance, see \cite{kutz2016multiresolution,callaham2019robust,MAULIK2023133852,maulik2020recurrent}), particularly for forecasting, extracting seasonal and long-term trends, and for flow-field recovery \cite{maulik2020probabilistic}. Each ``snapshot'' of data corresponds to an array of size 360 $\times$ 180 (based on a one-degree resolution). However, a mask is used to remove missing locations in the array that corresponds to the land area. Furthermore,  forecasts are performed for those coordinates which correspond to oceanic regions alone (and not inland bodies of water which are also masked). Unmasked data points are flattened to obtain a column vector for each snapshot of our training and test data.

The NOAA-SST data is available from October 22, 1981, to June 30, 2018 (i.e., 1,914 snapshots for the weekly averaged temperature). We utilize the period of October 22, 1981, to December 31, 1989 for training our model. From January 1990 onwards, the rest of the data is utilized for testing our various model deployments. This train-test split of the data set is commonly used in literature \cite{callaham2019robust} and the 8-year training period captures several important trends in the global sea surface temperature. Since this data set is produced by combining local and satellite temperature observations, it represents an attractive forecasting task for \emph{non-intrusive} data-driven methods without requiring the physical modeling of underlying processes.

Each snapshot of the SST, representing the system, is a vector in $\mathbb{R}^N$, where $N$ is the number of grid points. We calculate the mean:  $\bar{D} = \frac{1}{T} \sum_{t} D_t$, and hence define the mean-subtracted snapshots as:  $y_t = D_t - \bar{D}$. Proper Orthogonal Decomposition (POD) is utilized to identify a set of orthonormal basis vectors $(v_1, ..., v_M)$, where $M < N$, approximating spatial snapshots, such that $y_t \approx \sum_{j=1}^{M} a_j(t) v_j,\ t = 1, \ldots, T$. POD chooses the basis, $v_j$ , to minimize the residual with respect to the $L_2$ norm, $R = \sum_{t=1}^{T} ||y_t - \sum_{j=1}^{M} a_j(t)v_j ||^2$ Defining the snapshot matrix,  $S = [y_1| \ldots |y_T]$, the optimal basis is given by the $M$  eigenvectors of  $SS^T$  with largest eigenvalues. Afterward, the coefficients are found by orthogonal projection: $a(t) = \langle y_t, v \rangle.$, and models are trained to forecast future coefficients based on past ones. Predictions for coefficients are converted to physical space by adding the mean snapshot and a linear combination of coefficients and corresponding basis vectors, $\bar{D} + \sum_{j} \hat{a}_j v_j$. In this study, we use 5 POD modes to compress the original dataset for building an efficient time-series forecasting model.  We employ MinMaxZeroOne scaling to normalise the training input data POD modes to the reservoir into the range [0,1]. For the test POD modes, we reuse the same minimum and maximum feature values of the training modes to avoid data leakage.

We emphasize here that the proposed time-series forecasting technique makes predictions for the SST \emph{autoregressively} -- this is in comparison with past studies which have made windowed predictions (i.e., the ground truth input is always assumed to be observed). This mode of deployment is considerably more challenging as the trained model is expected to be stable to perturbations introduced by iterative prediction errors.

\section{Results}

In this section, we outline the results from our trained quantum reservoir models for forecasting the POD-compressed sea-surface temperature. We also compare the performance of the proposed approach with benchmark data-driven regression models for time-series data to establish any gains. We evaluate metrics from our forecasts of POD coefficients by reconstructing the state on the original latitude-longitude grid. Predictions are made for autoregressive rollouts over 300 predicted timesteps over which time-averaged metrics are obtained. Ensemble estimates are also provided for the top-5 models based on validation performance. Here, we define validation performance as the autoregressive rollout for 300 timesteps given a starting point within the first 127 snapshots.

\begin{figure}[h!]
    \centering
    \begin{subfigure}[b]{0.49\textwidth}
        \centering
        \includegraphics[width=\textwidth]{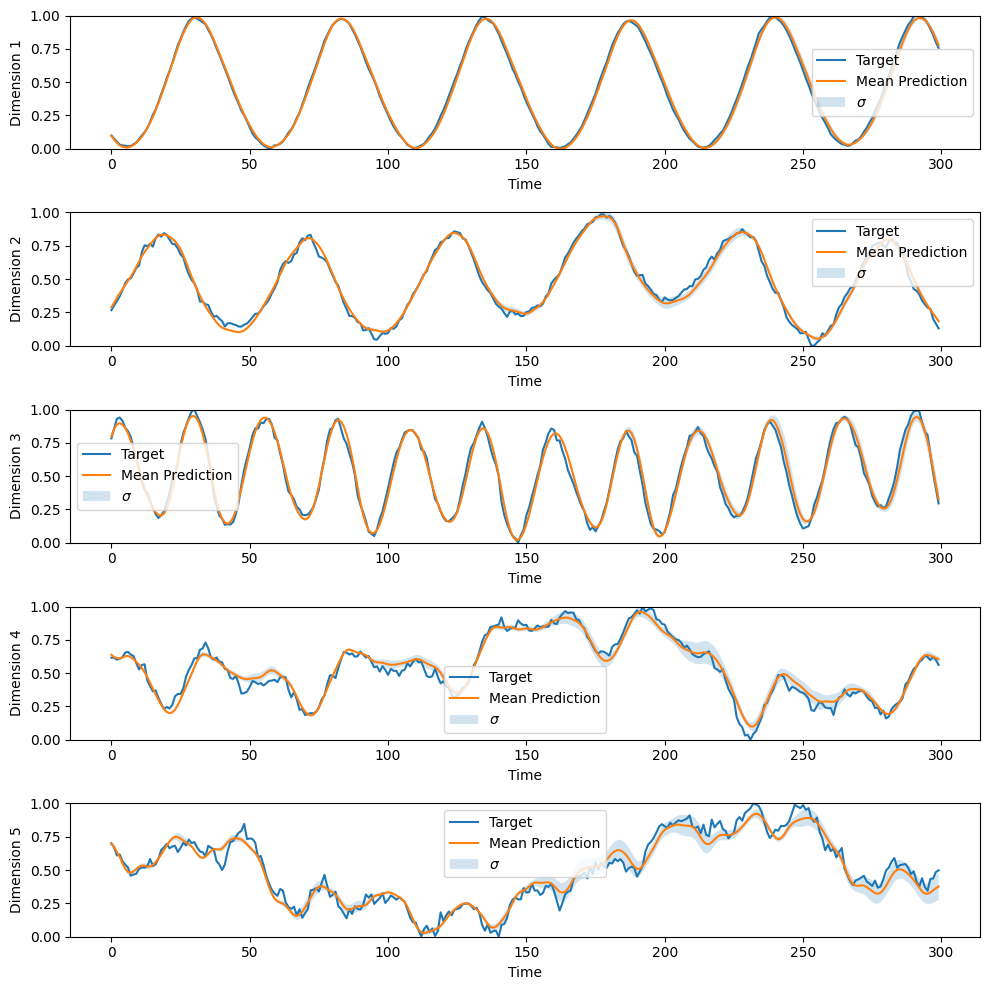}
        \caption{Training data predictions}
        \label{fig:HQRC-top5-train-dims}
    \end{subfigure}
    \begin{subfigure}[b]{0.49\textwidth}
        \centering
        \includegraphics[width=\textwidth]{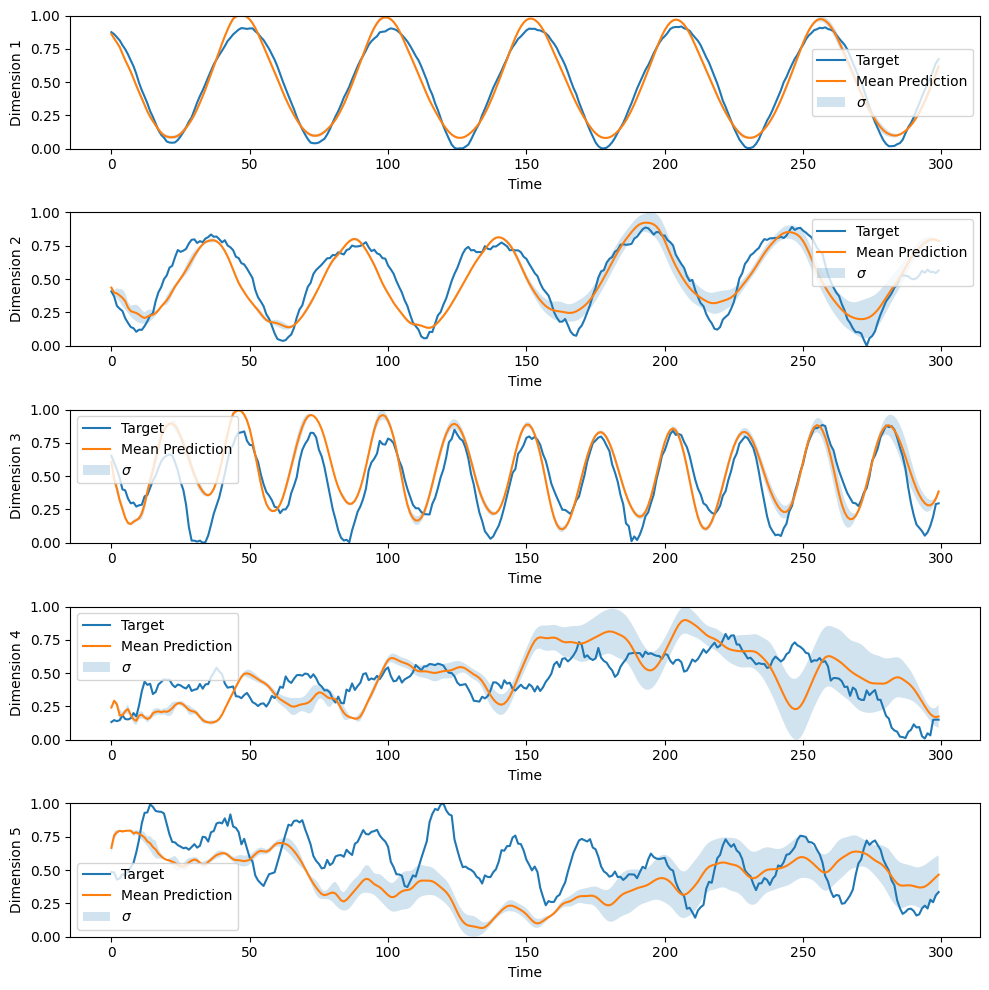}
        \caption{Test data predictions}
        \label{fig:HQRC-top5-test-dims}
    \end{subfigure}
    \caption{Average prediction and corresponding ensemble standard deviations for the top 5 HQRC models over train (left) and test (right) dataset predictions.}
    \label{fig:HQRC_dims}
\end{figure}

\begin{figure}[h!]
    \centering

    \mbox{
    \begin{subfigure}[b]{0.49\textwidth}
    \centering
    \includegraphics[width=\textwidth]{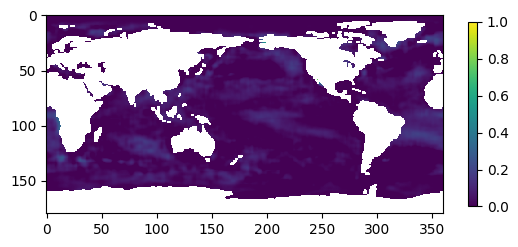}
    \caption{Best-cast reconstruction error for training data}    
    \end{subfigure}

    \begin{subfigure}[b]{0.49\textwidth}
    \centering
    \includegraphics[width=\textwidth]{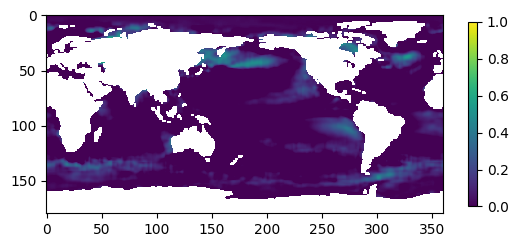}
    \caption{Best-cast reconstruction error for test data}    
    \end{subfigure}
    }    \\
    \mbox{

    \begin{subfigure}[b]{0.49\textwidth}
    \centering
    \includegraphics[width=\textwidth]{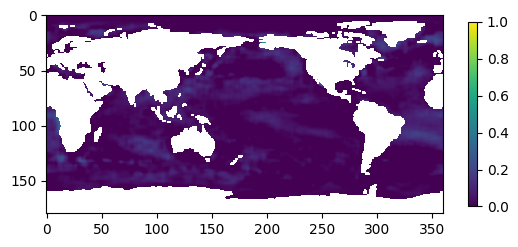}
    \caption{HQRC prediction error for training data}    
    \end{subfigure}

    \begin{subfigure}[b]{0.49\textwidth}
    \centering
    \includegraphics[width=\textwidth]{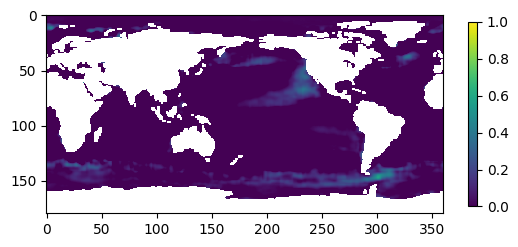}
    \caption{HQRC prediction error for test data}    
    \end{subfigure}
    }
    \caption{Best-case reconstruction error (i.e., using ground truth POD-coefficients) (top) and model-based prediction error of top 5 models of HQRC (bottom) for 300 averaged timesteps of predictions.}
    \label{fig:HQRC_errors}
\end{figure}

\begin{figure}[h!]
    \centering
    \begin{subfigure}[b]{0.49\textwidth}
        \centering
        \includegraphics[width=\textwidth]{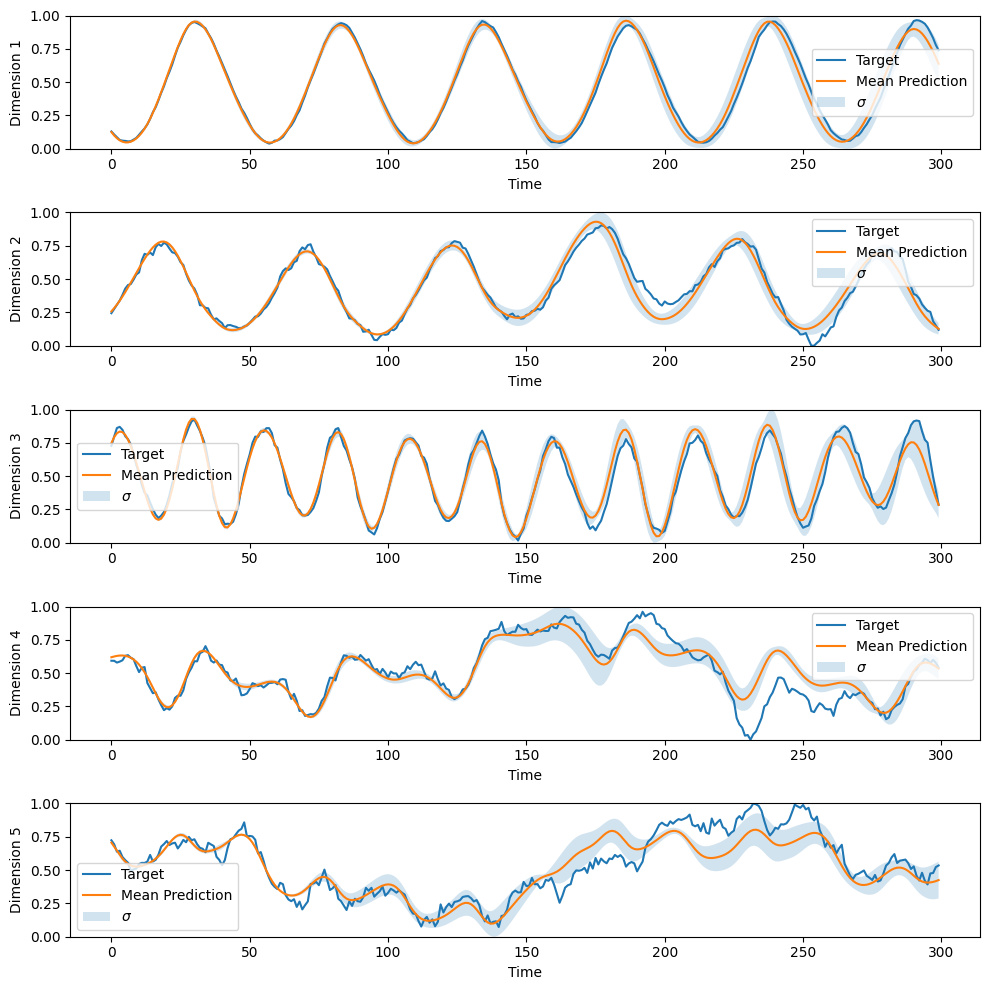}
        \caption{Training data predictions}
        \label{fig:GRU-top5-train-dims}
    \end{subfigure}
    \begin{subfigure}[b]{0.49\textwidth}
        \centering
        \includegraphics[width=\textwidth]{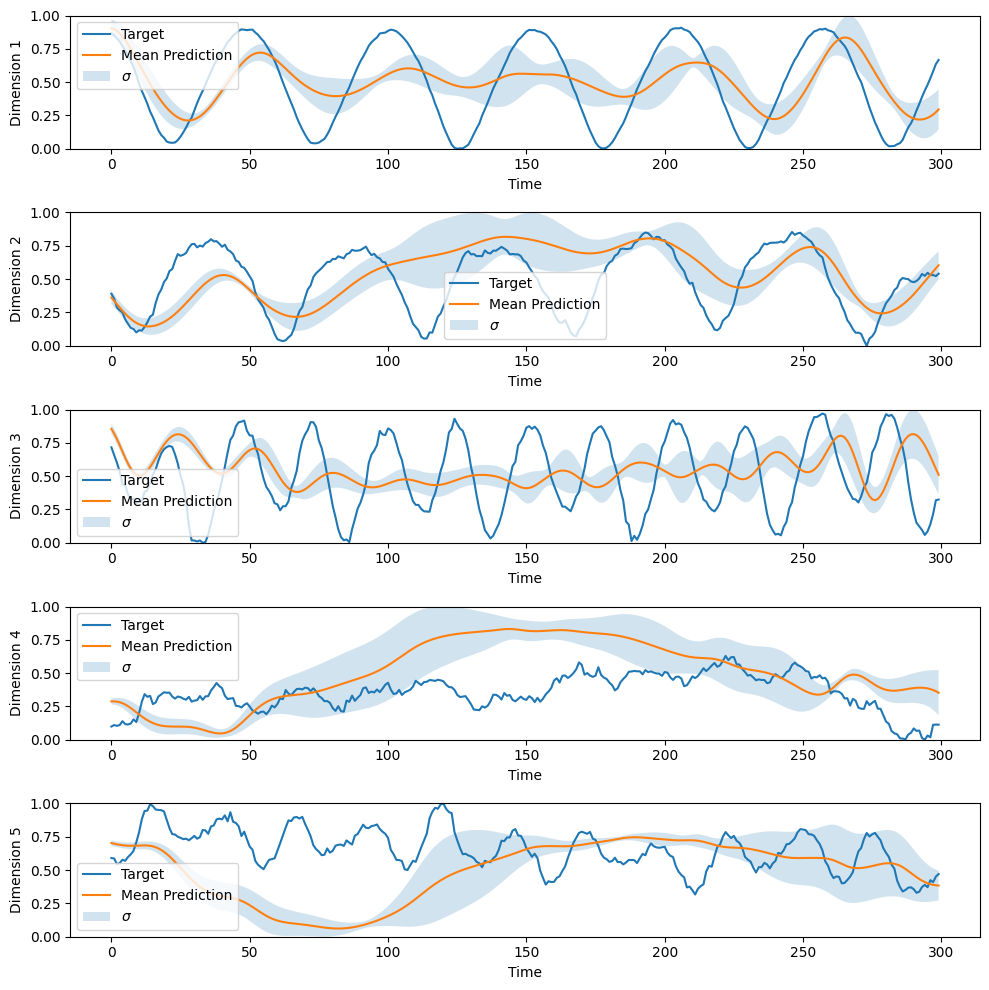}
        \caption{Test data predictions}
        \label{fig:GRU-top5-test-dims}
    \end{subfigure}
    \caption{Average prediction and corresponding ensemble standard deviations for the top 5 GRU models over train (left) and test (right) dataset predictions.}
    \label{fig:GRU_dims}
\end{figure}

\begin{figure}[h!]
    \centering
    \mbox{

    \begin{subfigure}[b]{0.49\textwidth}
    \centering
    \includegraphics[width=\textwidth]{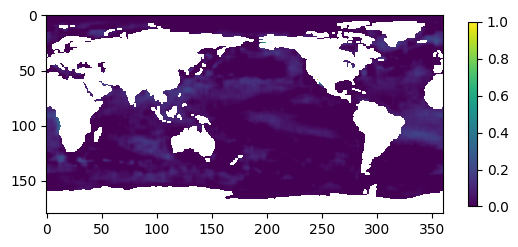}
    \caption{Prediction error for training data}    
    \end{subfigure}

    \begin{subfigure}[b]{0.49\textwidth}
    \centering
    \includegraphics[width=\textwidth]{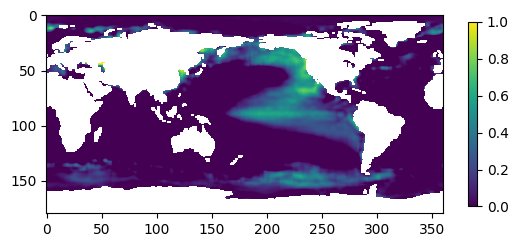}
    \caption{Prediction error for test data}    
    \end{subfigure}
    }
    \caption{Model-based prediction error of top 5 GRU models for 300 averaged timesteps of predictions.}
    \label{fig:GRU_errors}
\end{figure}

\begin{figure}[h!]
    \centering
    \begin{subfigure}[b]{0.49\textwidth}
        \centering
        \includegraphics[width=\textwidth]{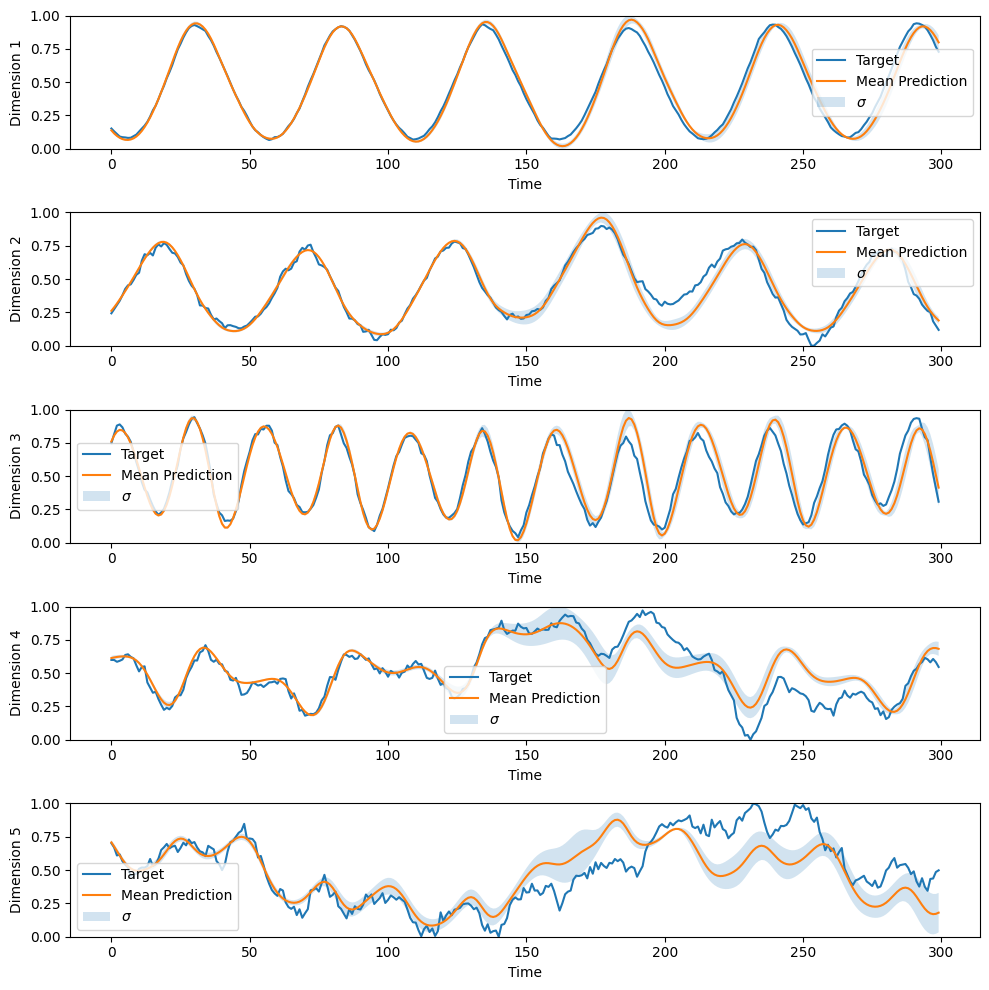}
        \caption{Training data predictions}
        \label{fig:LSTM-top5-train-dims}
    \end{subfigure}
    \begin{subfigure}[b]{0.49\textwidth}
        \centering
        \includegraphics[width=\textwidth]{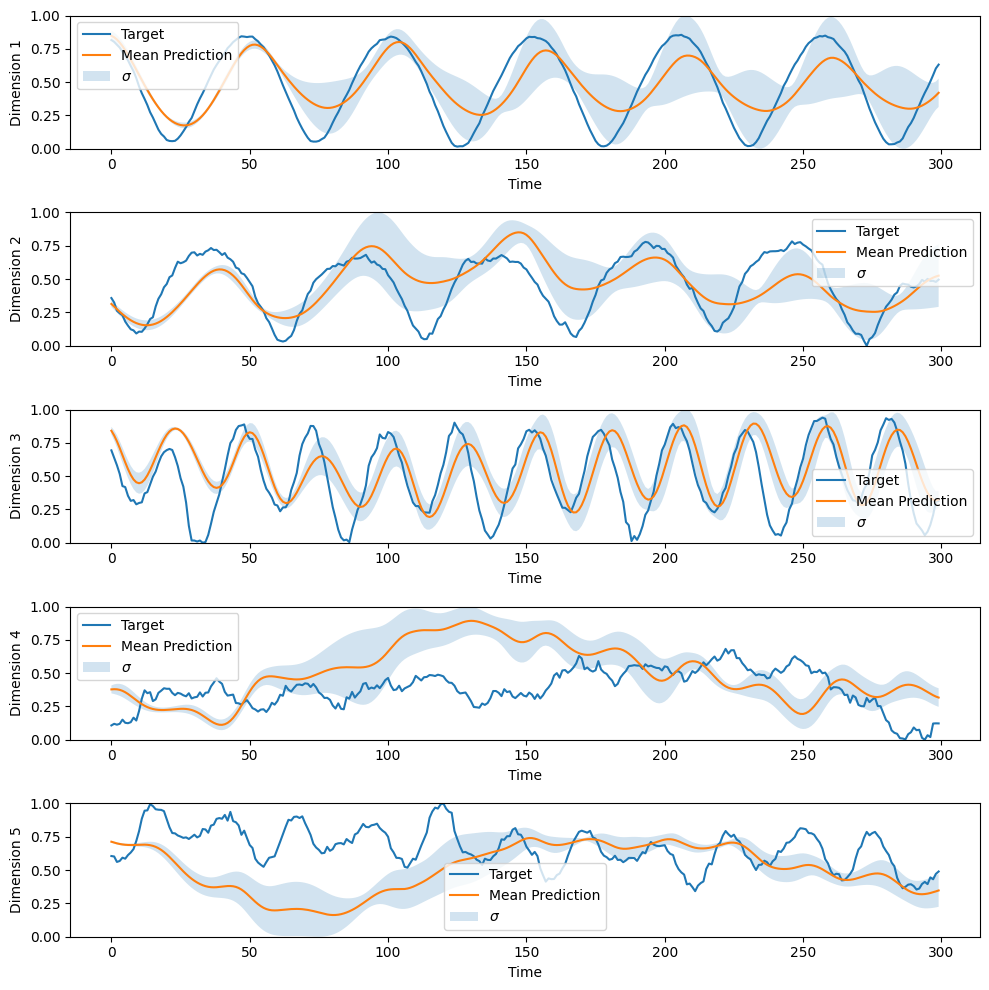}
        \caption{Test data predictions}
        \label{fig:LSTM-top5-test-dims}
    \end{subfigure}
    \caption{Average prediction and corresponding ensemble standard deviations for the top 5 LSTM models over train (left) and test (right) dataset predictions.}
    \label{fig:LSTM_dims}
\end{figure}

\begin{figure}[h!]
    \centering
    \mbox{

    \begin{subfigure}[b]{0.49\textwidth}
    \centering
    \includegraphics[width=\textwidth]{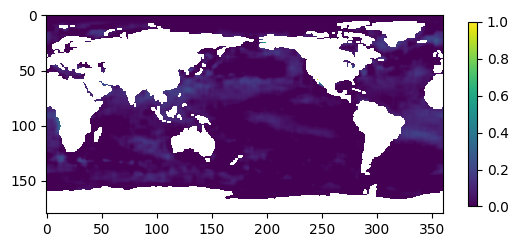}
    \caption{Prediction error for training data}    
    \end{subfigure}

    \begin{subfigure}[b]{0.49\textwidth}
    \centering
    \includegraphics[width=\textwidth]{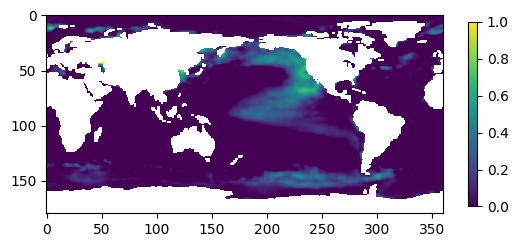}
    \caption{Prediction error for test data}    
    \end{subfigure}
    }
    \caption{Model-based prediction error of top 5 LSTM models for 300 averaged timesteps of predictions.}
    \label{fig:LSTM_errors}
\end{figure}

\begin{figure}[h!]
    \centering
    \begin{subfigure}[b]{0.49\textwidth}
        \centering
        \includegraphics[width=\textwidth]{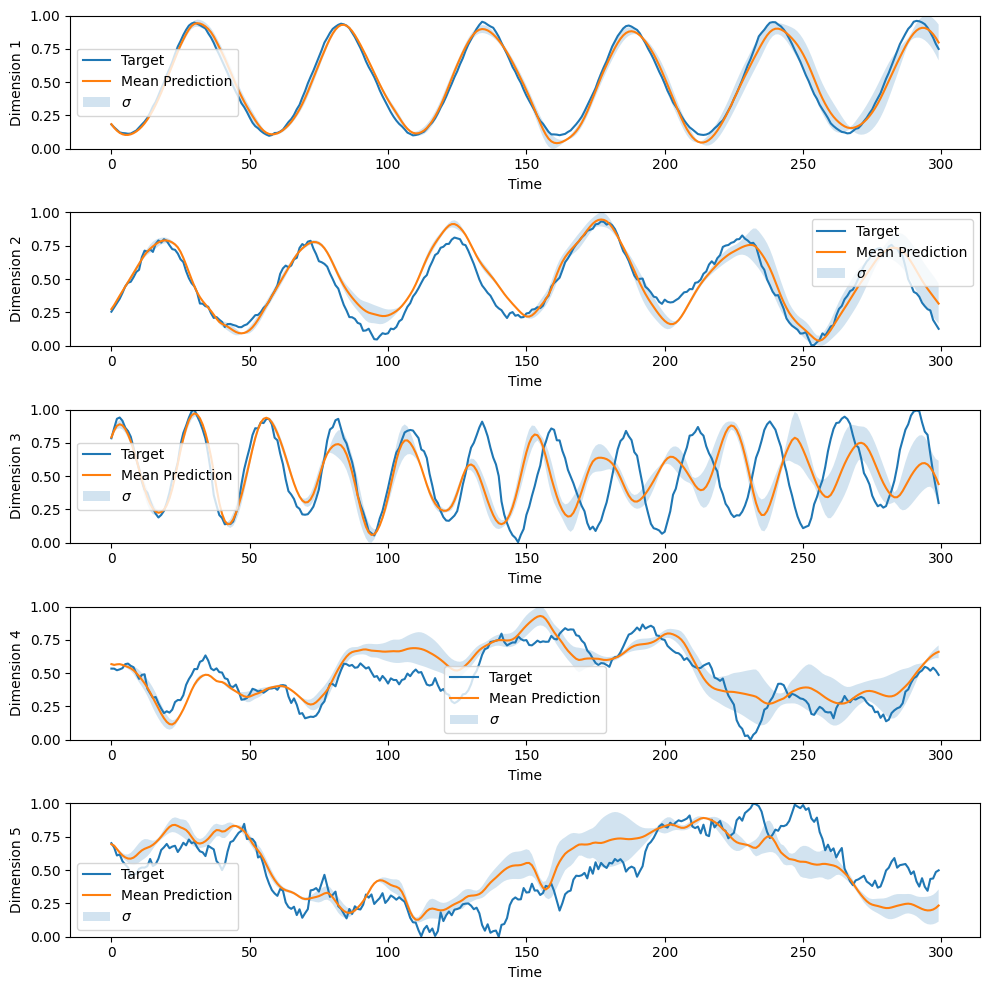}
        \caption{Training data predictions}
        \label{fig:ESN-top5-train-dims}
    \end{subfigure}
    \begin{subfigure}[b]{0.49\textwidth}
        \centering
        \includegraphics[width=\textwidth]{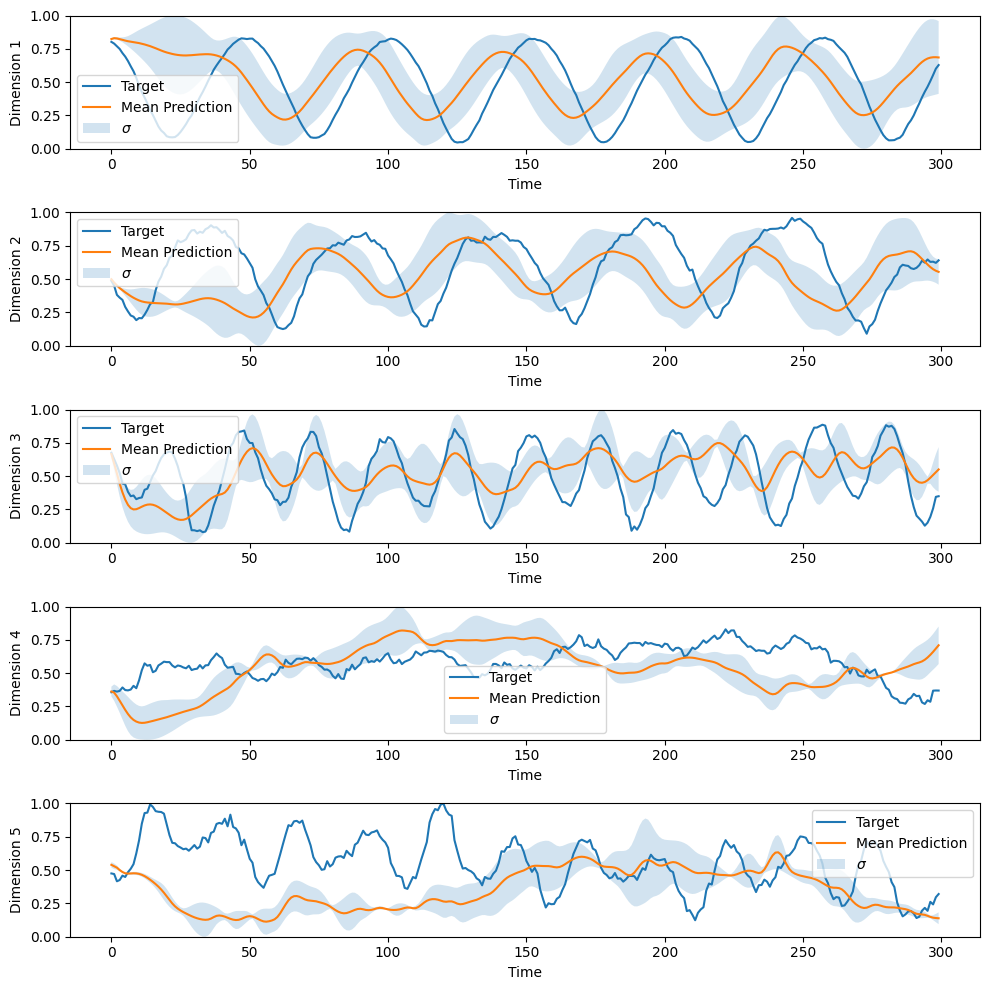}
        \caption{Test data predictions}
        \label{fig:ESN-top5-test-dims}
    \end{subfigure}
    \caption{Average prediction and corresponding ensemble standard deviations for the top 5 ESN models over train (left) and test (right) dataset predictions.}
    \label{fig:ESN_dims}
\end{figure}

\begin{figure}[h!]
    \centering
    \mbox{

    \begin{subfigure}[b]{0.49\textwidth}
    \centering
    \includegraphics[width=\textwidth]{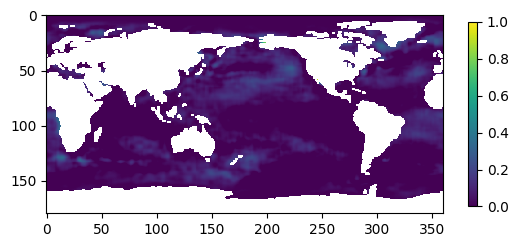}
    \caption{Prediction error for training data}    
    \end{subfigure}

    \begin{subfigure}[b]{0.49\textwidth}
    \centering
    \includegraphics[width=\textwidth]{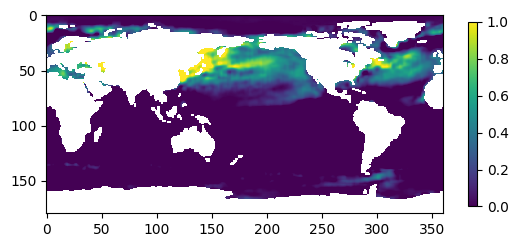}
    \caption{Prediction error for test data}    
    \end{subfigure}
    }
    \caption{Model-based prediction error of top 5 ESN models for 300 averaged timesteps of predictions.}
    \label{fig:ESN_errors}
\end{figure}

We average the predictions from five different models, each with the best performance on the training data. These models were trained after tuning a set of hyperparameters, the results of which can be found in the Appendix. To ensure fairness, we spent approximately 10 hours optimizing the hyperparameters for each model using an Intel(R) Core(TM) i7-10510U CPU @ 1.80GHz 2.30 GHz. For consistency, we predicted 300 weeks (time steps) from a randomly selected starting point during the test duration. In all the figures displayed, we have started from the 116th week.
Figure \ref{fig:HQRC_dims}, \ref{fig:GRU_dims}, \ref{fig:LSTM_dims} and \ref{fig:ESN_dims} shows the target and average prediction and the corresponding standard deviation (shaded) over the ensemble of the top 5 best-performing models on training and test data modal coefficients. Figure \ref{fig:HQRC_errors}, \ref{fig:GRU_errors}, \ref{fig:LSTM_errors} and \ref{fig:ESN_errors} show the reconstruction and prediction error grids for training and test data for the last prediction week. The land region is masked. The RMSE errors over the entire grid and for the East Pacific region (corresponding to latitudes and longitudes in the region between -10 to +10 degrees latitude and 200-250 degrees longitude) for test and training data, averaged over 300 predicted timesteps, are all shown in Table \ref{tab:model_comparison} along with computational performance metrics like training time, memory and number of parameters used. It is to be noted that the lowest-possible reconstruction error for training data is 0.458, and for test data, it is 0.491, so expecting a lower RMSE is impractical. This error corresponds to that left over from the process of projecting the original state on to reduced-set of POD basis vectors and is irreducible.

It is to be noted that the models with the best performance in training data need not be the ones with the best performance on test data as well (and were often observed to be different). This is usually attributed to overfitting and the usual strategies for mitigation have been applied here (such as early stopping on held-out validation data). It is observed that the HQRC method clearly outperforms GRUs, LSTMs and ESNs in terms of the RMSE error metrics. Moreover, computationally, it is significantly less expensive than LSTMs and GRUs to train and deploy, as it takes less time to train and requires fewer trainable parameters. However, the classical reservoir-based model i.e ESNs, although computationally cheaper to train, perform worse than the other methods. The HQRC model thus proves to be an effective balance of the efficiency of the ESN and the learning abilities of more complex neural network based models for forecasting. We go into the specific details of each model's performance in the following.

\noindent \textbf{HQRC:} The model is able to accurately capture the trend for all dimensions of the training data; It able to capture the trend for the test data fairly well for the first three dimensions, it is even able to capture the large scale trends for the last two dimensions of the test data, but it is unable to capture the small scale fluctuations well.
The model is particularly stable in its performance, and the shaded region, i.e. the standard deviation, only becomes prominent for the later timesteps for the test data. We perform several ablation studies for the HQRC, including the influence of the input-window length, the number of reservoirs, the effect of compression (and consequently dimensionality), as well as the sensitivity to initial condition perturbations in the Appendix. 

\noindent \textbf{GRU:} While GRU performs fairly well for training data, deviations are observed over the duration of the rollout. It can also be observed that the standard deviation between the models increases as the prediction proceeds autoregressively. For test data, the prediction becomes flat for all the dimensions, along with a huge standard deviation, indicating a disagreement between the models (this is another commonly observed mode of failure for autoregressively deployed time-series forecasting models). As observed from the prediction reconstruction for test data (Figure \ref{fig:GRU_errors}), a significant fraction of the domain shows high errors.

\noindent \textbf{LSTM:} LSTMs tend to perform worse on training data, but perform better on test data in comparison with GRUs. This is likely because GRU deployments overfit on training data. However, a similar gradual convergence to a fixed-point is obtained during autoregressively deployments for the LSTM as well (although this is less pronounced than the GRU). For a fair comparison of LSTMs and GRUs, we utilize approximately the same input window length. We note that for both LSTMs and GRUs this window length is given by $SL + HSPL/4$ where $SL$ is the sequence length and $HSPL$ is the hidden state propagation length. The former refers to the duration in time for which backpropagation is performed before encoding the input sequence and the latter refers to number of times the hidden state is recursively updated before the prediction commences. These values are set according to the parameters chosen in Vlachas et al., \cite{vlachas2020backpropagation}.  Note that the prediction length (PL) is set to 1 for an autoregressive rollout.


\noindent \textbf{ESN:} Finally, we assess the performance of ESNs which are very similar in terms of computational expense and number of parameters to the HQRC. We observe large errors during testing when deploying the ESN, despite good performance during training (for the top 5 models by validation error) - which suggests a tendency to overfit. We note that the issue of flatlining during test deployments is not observed unlike the GRU and LSTM models. Specific details of the best hyperparameters for the QRC and other benchmark models are available in the Appendix. 

\begin{table}[h!]
    \centering
    \begin{tabular}{|l|c|c|c|c|}
        \toprule
        \textbf{Performance} & \textbf{HQRC} & \textbf{LSTM} & \textbf{GRU} & \textbf{ESN} \\
        \midrule
        \textbf{Average Training Time (s)} & 22.56 & 770.86 & 788.335 & \textbf{0.135} \\
        \textbf{Average Memory Taken (MB)} & 107.94 & 1179.01 & 989.86 & \textbf{101.99} \\
        \textbf{Number of Parameters} & \textbf{755-3005} & 27925-48066005 & 21045-36052005 & 200-5000 \\
        \textbf{RMSE Error (Train)} & \textbf{0.473} & 0.528 & 0.511 & 0.596 \\
        \textbf{RMSE Error in East Pacific Region (Train)} & \textbf{0.545} & 0.732 & 0.678 & 0.780 \\
        \textbf{RMSE Error (Test)} & \textbf{0.639} & 1.095 & 1.374 & 1.770 \\
        \textbf{RMSE Error in East Pacific Region (Test)} & \textbf{0.873} & 1.295 & 1.466 & 1.581 \\
        \bottomrule
    \end{tabular}
    \caption{Comparison of Different Models: Our experiments show that the HQRC approach obtains competitive forecast accuracy at reasonable training time and memory consumption. Note that these errors are computed over the course of a 300 step rollout and averaged across the top-5 models obtained during training.}
    \label{tab:model_comparison}
\end{table}

\section{Conclusion}

In this paper, we have introduced a novel time-series forecasting technique for compressed representations of dynamical systems using higher-order quantum reservoir computing. Our approach relies on the quantum embedding of the state-space into a density matrix that may be evolved by unitary operations performed by multiple quantum reservoirs. This leads to the emergence of nonlinearity from solely linear operations and is viable for forecasting low-dimensional representations of complex nonlinear dynamical systems. Our proposed approach is efficient in terms of training time and memory consumption and more accurate when compared to classical deep learning based methods for forecasting regression tasks. We demonstrate this on the task of forecasting a real-world complex dynamical system benchmark given by the NOAA optimum interpolation sea-surface temperature dataset. However, we also observe limitations of our method when increasing the dimensionality of the state-space which motivates future studies with respect to optimal selection of state-space (i.e., modal) components from our time-series data before the construction of the density matrix. Furthermore, future work may also benefit from examinations of HQRC performance on multiple modalities of data - for instance when time-series from different modalities of sensing may be incorporated into the state-vector evolution.

\section*{Acknowledgements}

This research used resources of the Argonne Leadership Computing Facility, which is a U.S. Department of Energy Office of Science User Facility operated under contract DE-AC02-06CH11357. RM acknowledges funding support from ASCR for DOE-FOA-2493 ``Data-intensive scientific machine learning''.


\bibliographystyle{unsrt}  
\bibliography{references}  

\appendix

\section*{Appendix}
For all the models, the NRMSE for the coefficients is averaged over 3 test cases selected randomly as starting points. Testing indices (Week from which the prediction starts $\in \{116,40, 66\}$) for over 300 timesteps. The NRMSE is defined as $\sqrt{\langle \frac{(Y_{\text{true}} - Y_{\text{pred}})^2}{\sigma^2} \rangle}$ where $\langle . \rangle$ denotes average over all timesteps.

\begin{table}[h!]
\centering
\begin{tabular}{|l|l|}
\hline
\textbf{Hyperparameter} & \textbf{Value(s)} \\ \hline
Number of reservoirs ($N_{qrc}$) & 5 \\ \hline
Coupling coefficient of the Ising Hamiltonian & 2.0 \\ \hline
Interval of time for evolution $\tau$ & 4 \\ \hline
Number of qubits for each reservoir $N_l$ & 6 \\ \hline
Dynamics length (DL) for washout phase & 40 \\ \hline
Iterative Prediction length (PL) & 300 \\ \hline
Number of Virtual nodes $V$ & \{5, 10, 15, 20\} \\ \hline
Connection strength $\alpha$ & \{0.3, 0.4, 0.6, 0.7, 0.8, 0.9\} \\ \hline
Ridge Parameter $\beta$ & \{$10^{-3}$, $10^{-4}$, $10^{-5}$, $10^{-6}$, $10^{-7}$\} \\ \hline
\end{tabular}
\caption{List of hyperparameters searched over for HQRC and their values}
\label{tab:HQRC-params}
\end{table}

\begin{table}[h!]
    \centering
    \begin{tabular}{|c|c|}
        \hline
        \textbf{Model} & \textbf{RMNSE} \\
        \hline
        HQRC-V=10-alpha=0.5-beta=1e-07 & 0.1573 \\
        HQRC-V=10-alpha=0.4-beta=1e-06 & 0.1759 \\
        HQRC-V=10-alpha=0.4-beta=1e-05 & 0.1864 \\
        HQRC-V=10-alpha=0.3-beta=1e-05 & 0.1941 \\
        HQRC-V=10-alpha=0.5-beta=1e-06 & 0.1995 \\
        \hline
    \end{tabular}
    \caption{Top 5 HQRC Models and their RMNSE Values over the training data coefficients}
    \label{tab:HQRC_top5}
\end{table}

\begin{table}[h!]
\centering
\begin{tabular}{|l|l|}
\hline
\textbf{Hyperparameter} & \textbf{Value(s)} \\ \hline
Batch size & 32 \\ \hline
Number of retraining rounds & 5 \\ \hline
Maximum epochs & 100 \\ \hline
Overfitting patience & 20 \\ \hline
Learning rate & 0.001 \\ \hline
Training validation split ratio & 0.7 \\ \hline
Iterative Prediction length (PL) & 300 \\ \hline
Size of layer & \{80, 120, 250, 500, 1000, 2000\} \\ \hline
Number of layers & \{1, 2\} \\ \hline
Sequence length (for training) & \{16, 20\} \\ \hline
Prediction length (kept same as SL) & \{16, 20\} \\ \hline
Hidden state propagation length & \{50, 60, 70\} \\ \hline
\end{tabular}
\caption{List of hyperparameters searched over for GRU and their values}
\label{tab:GRU-params}
\end{table}

\begin{table}[h!]
    \centering
    \begin{tabular}{|c|c|}
        \hline
        \textbf{Model} & \textbf{RMNSE} \\
        \hline
        GRU-250-(2)-HSPL-70-SL-16 & 0.2735 \\
        GRU-80-(1)-HSPL-70-SL-16 & 0.3024 \\
        GRU-80-(1)-HSPL-60-SL-20 & 0.3162 \\
        GRU-120-(2)-HSPL-70-SL-16 & 0.3272 \\
        GRU-120-(1)-HSPL-70-SL-16 & 0.3409 \\
        \hline
    \end{tabular}
    \caption{Top 5 GRU Models and their RMNSE Values for training data coefficients}
    \label{tab:GRU_top5}
\end{table}

\begin{table}[h!]
\centering
\begin{tabular}{|l|l|}
\hline
\textbf{Hyperparameter} & \textbf{Value(s)} \\ \hline
Batch size & 32 \\ \hline
Number of retraining rounds & 5 \\ \hline
Maximum epochs & 100 \\ \hline
Overfitting patience & 20 \\ \hline
Learning rate & 0.001 \\ \hline
Training validation split ratio & 0.7 \\ \hline
Iterative Prediction length (PL) & 300 \\ \hline
Size of layer & \{80, 120, 250, 500, 1000, 2000\} \\ \hline
Number of layers & \{1, 2\} \\ \hline
Sequence length (for training) & \{16, 20\} \\ \hline
Prediction length (kept same as SL) & \{16, 20\} \\ \hline
Hidden state propagation length & \{50, 60, 70\} \\ \hline
\end{tabular}
\caption{List of hyperparameters searched over for LSTM and their values}
\label{tab:LSTM-params}
\end{table}

\begin{table}[h!]
    \centering
    \begin{tabular}{|c|c|}
        \hline
        \textbf{Model} & \textbf{RMNSE} \\
        \hline
        LSTM-80-(1)-HSPL-70-SL-16 & 0.2313 \\
        LSTM-120-(1)-HSPL-70-SL-16 & 0.2922 \\
        LSTM-2000-(1)-HSPL-50-SL-20 & 0.3230 \\
        LSTM-120-(1)-HSPL-50-SL-16 & 0.3241 \\
        LSTM-250-(1)-HSPL-70-SL-16 & 0.3513 \\
        \hline
    \end{tabular}
    \caption{Top 5 LSTM Models and their RMNSE Values for training data coefficients}
    \label{tab:LSTM_top5}
\end{table}

\begin{table}[h!]
\centering
\begin{tabular}{|l|l|}
\hline
\textbf{Hyperparameter} & \textbf{Value(s)} \\ \hline
Degree of nodes & 10 \\ \hline
Radius & 0.9 \\ \hline
Learning rate & 0.001 \\ \hline
Number of epochs & 1000000 \\ \hline
Dynamics length (DL) & 40 \\ \hline
Iterative Prediction length (PL) & 300 \\ \hline
Number of Nodes (UNITS) & \{40, 50, 60, 70, 80, 100, 120, 150, 200, 300, 500, 1000\} \\ \hline
Ridge Parameter $\beta$ & \{$10^{-4}$, $10^{-5}$, $10^{-6}$, $10^{-7}$\} \\ \hline
\end{tabular}
\caption{List of hyperparameters searched over for ESN and their values}
\label{tab:ESN-param}
\end{table}

\begin{table}[h!]
    \centering
    \begin{tabular}{|c|c|}
        \hline
        \textbf{Model} & \textbf{RMNSE} \\
        \hline
        ESN-60-beta-1e-05 & 0.5201 \\
        ESN-60-beta-0.0001 & 0.5268 \\
        ESN-60-beta-1e-06 & 0.5280 \\
        ESN-60-beta-1e-07 & 0.5289 \\
        ESN-100-beta-0.0001 & 0.7686 \\
        \hline
    \end{tabular}
    \caption{Top 5 ESN Models and their RMNSE Values for training data coefficients}
    \label{tab:ESN_top5}
\end{table}

\begin{table}[h!]
    \centering
    \begin{tabular}{|c|c|}
        \hline
        \textbf{Model} & \textbf{RMNSE} \\
        \hline
        HQRC-V=1-alpha=0.5-beta=1e-07-NQRC-10 & 1.2402 \\
        HQRC-V=5-alpha=0.5-beta=1e-07-NQRC-10 & 1.5721 \\
        HQRC-V=5-alpha=0.5-beta=1e-07-NQRC-20 & 1.5939 \\
        HQRC-V=10-alpha=0.5-beta=1e-07-NQRC-10 & 1.6087 \\
        HQRC-V=10-alpha=0.5-beta=1e-07-NQRC-20 & 1.8560 \\
        \hline
    \end{tabular}
    \caption{Top 5 HQRC Models with 10 Modes and their RMNSE Values for training data coefficients}
    \label{tab:HQRC_top5_10modes}
\end{table}

\subsection*{Effect of HQRC washout length}

In the above experiments, We used 40 warming-up steps, i.e. 40 timesteps were fed into the Quantum reservoir model before starting the prediction for 300 steps. This was done to ensure that the reservoir state $z$ captured the trend of the time series and to wash out or forget the bias of the initial state \cite{tran2020higherorder}. We wanted to evaluate the performance of the QRC model for different dynamics lengths (or washout steps). Table \ref{tab:HQRC-DL_comparison} shows the RMSE results for different dynamics lengths, keeping other hyperparameters constant (corresponding to the model with the best training data performance). As expected, the performance seems to improve as we increase the dynamics length, this is because the reservoir state is better able to washout the effect of the initial state and capture the dynamics of the data. 
\begin{table}[h!]
    \centering
    \begin{tabular}{|l|c|c|c|}
        \toprule
        \textbf{Performance} & \textbf{DL-20} & \textbf{DL-30} & \textbf{DL-40} \\
        \midrule
        \textbf{RMSE Error (Train)} & 0.705 & 0.525 & 0.468 \\
        \textbf{RMSE Error in Pacific Region (Train)} & 1.68 & 0.972 & 0.558 \\
        \textbf{RMSE Error (Test)} & 0.789 & 0.679 & 0.669 \\
        \textbf{RMSE Error in Pacific Region (Test)} & 1.151 & 1.154 & 1.138 \\
        \bottomrule
    \end{tabular}
    \caption{Comparison for Different Dynamics Length (DL) for HQRC with V=10, $\alpha=0.5, \beta=10^{-7}$}
    \label{tab:HQRC-DL_comparison}
\end{table}


\subsection*{Sensitivity to initial condition perturbations}

\begin{figure}[h!]
  \centering
  \includegraphics[width=0.8\textwidth]{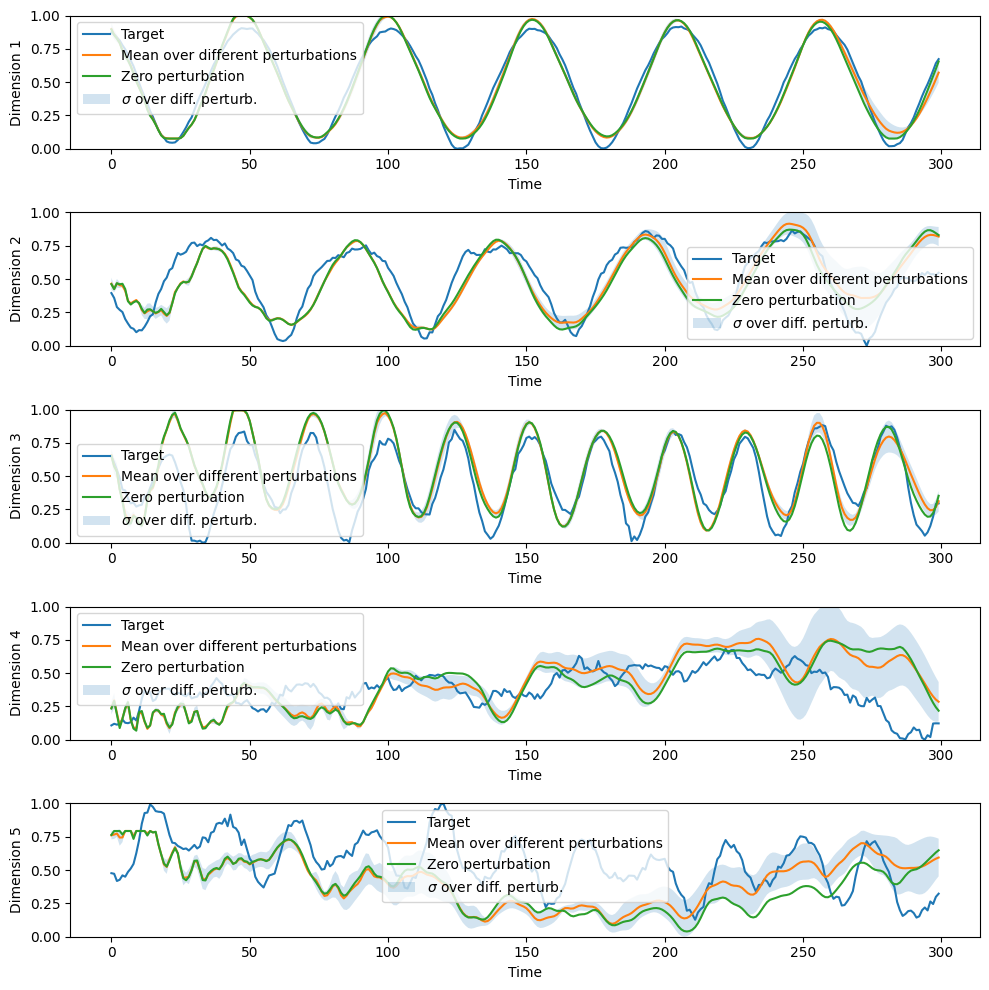}
  \caption{HQRC model with slight perturbations added at the first timestep.}
  \label{fig:perturbations}
\end{figure}

Figure \ref{fig:perturbations} Shows predictions when slight perturbations added at the first timestep. The model is trained with V=10, $\alpha = 0.5$, $\beta = 10^-7$ and rest is same as above. The mean over perturbations and the zero perturbation predictions tend to coincide for the most part for the first three dimensions, indicating the stability of the model. For the last two dimensions, the standard deviation over different perturbations tends to increase with time.

\subsection*{The effect of modal compression}

Instead of keeping just 5 modes, we decided to keep 10 modes to see to if the performance is affected. Figure \ref{fig:modes-hqrc-test-dims} shows the average prediction over the top 5 models based on training data (see appendix), for the 300 timesteps for the test data. We note that the predictions fluctuate a lot, and the standard deviation between the models is very high.

\begin{figure}[h!]
    \mbox{
    \begin{subfigure}[b]{0.49\textwidth}
    \centering
    \includegraphics[width=\textwidth]{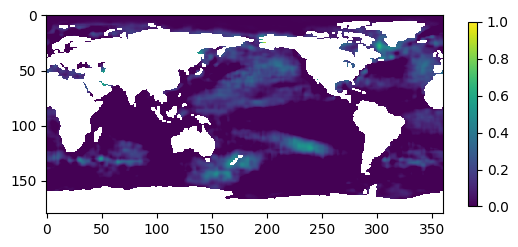}
    \caption{Prediction error for training data}    
    \end{subfigure}

    \begin{subfigure}[b]{0.49\textwidth}
    \centering
    \includegraphics[width=\textwidth]{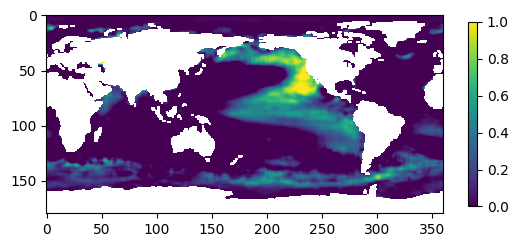}
    \caption{Prediction error for test data}    
    \end{subfigure}
    }
    \caption{Model-based prediction error of top 5 HQRC models (using 10 modes for compression) for 300 averaged timesteps of predictions.}
    \label{fig:modes-hqrc-recon-test}
\end{figure}

The final RMSE error for the training data over all 300 predicted timesteps (between $116^{th}$ and $415^{th}$ weeks) is 1.220, compared with the corresponding default reconstruction RMSE of 0.458. The RMSE restricted to the East pacific region for the HQRC prediction is 1.414.
The RMSE between $116^{th}$ and $415^{th}$ weeks of the test data is 1.095 and in the East pacific region is 1.305.
Figure \ref{fig:modes-hqrc-recon-test} shows the reconstruction error grid from the 10 predicted mode coefficients.

\begin{figure}[h!]
    \centering
    \includegraphics[width=0.9\linewidth]{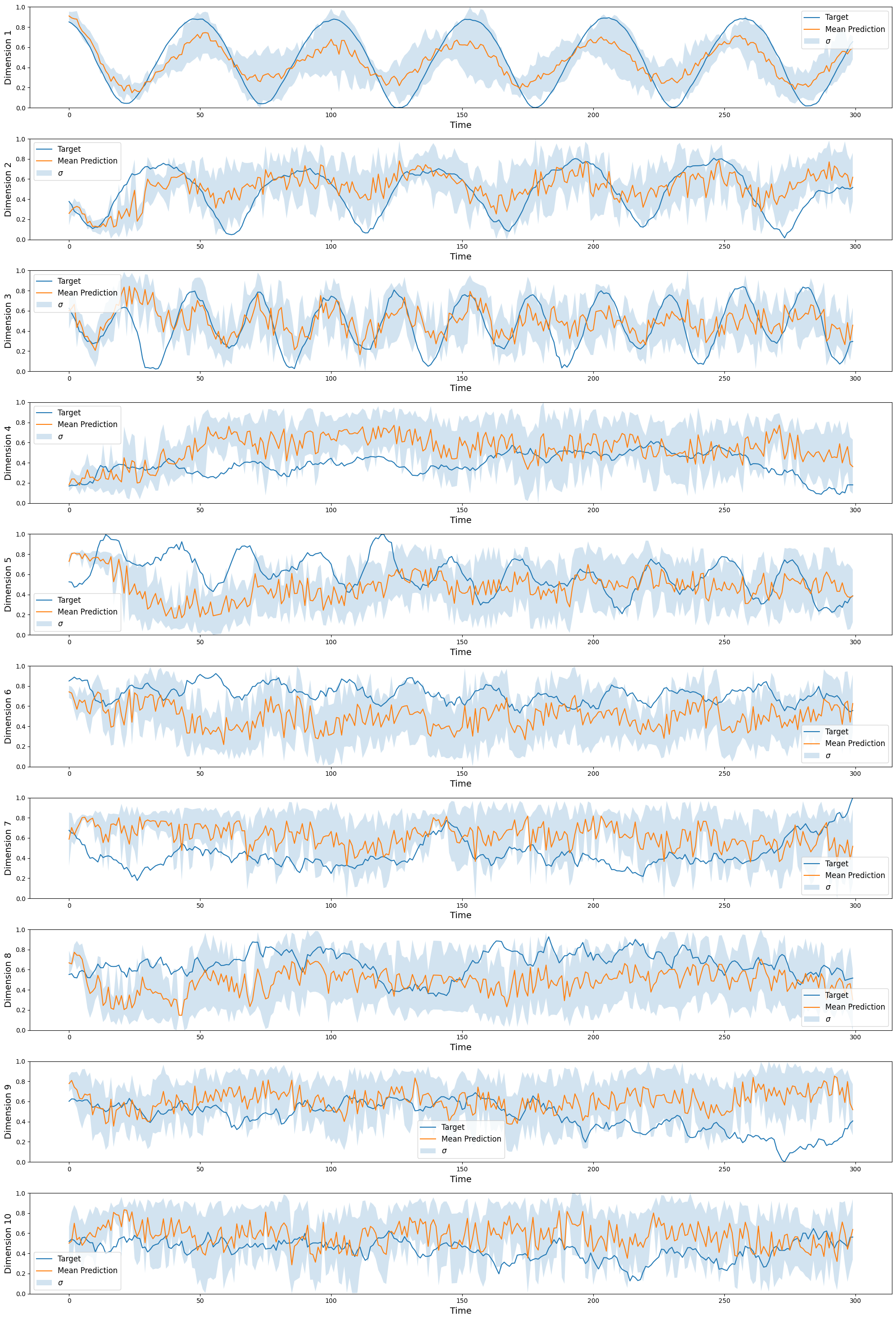}
    \caption{10 modes of SST data prediction using HQRC, for top 5 models of HQRC}
    \label{fig:modes-hqrc-test-dims}
\end{figure}

\subsection*{Results for different Qubits/Reservoirs}
\begin{figure}[h!]
    \centering
    \begin{subfigure}[]{0.48\textwidth}
        \includegraphics[width=\textwidth]{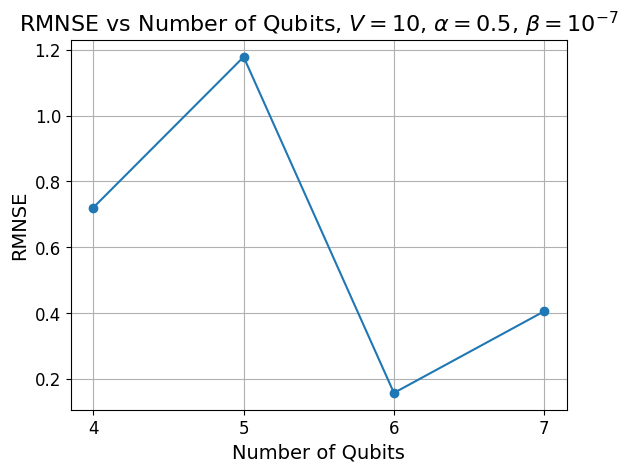}
        \caption{RMNSE error of modes vs Number of qubits}
        \label{fig:params-nqubits}
    \end{subfigure}
    \hfill
    \begin{subfigure}[]{0.48\textwidth}
        \includegraphics[width=\textwidth]{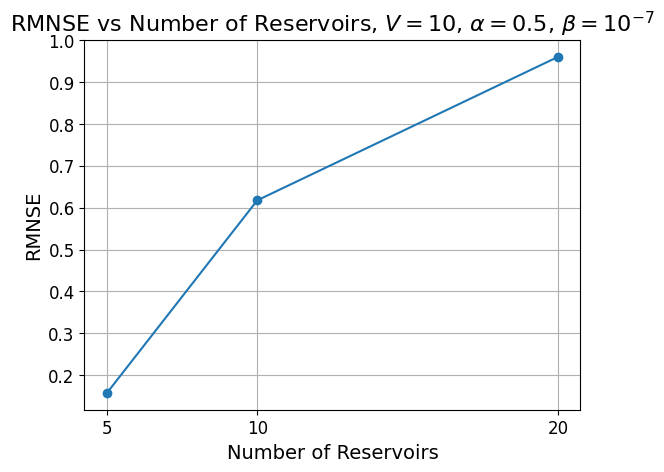}
        \caption{RMNSE error of modes vs Number of Reservoirs}
        \label{fig:params-nreservoirs}
    \end{subfigure}
    \begin{subfigure}[]{0.48\textwidth}
        \includegraphics[width=\textwidth]{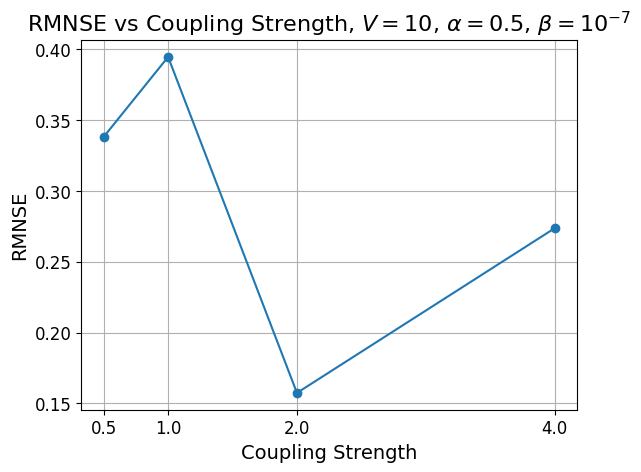}
        \caption{RMNSE error of modes vs Coupling Strength of Ising Hamiltonian}
        \label{fig:params-coupling}
    \end{subfigure}
    \hfill
    \begin{subfigure}[]{0.48\textwidth}
        \includegraphics[width=\textwidth]{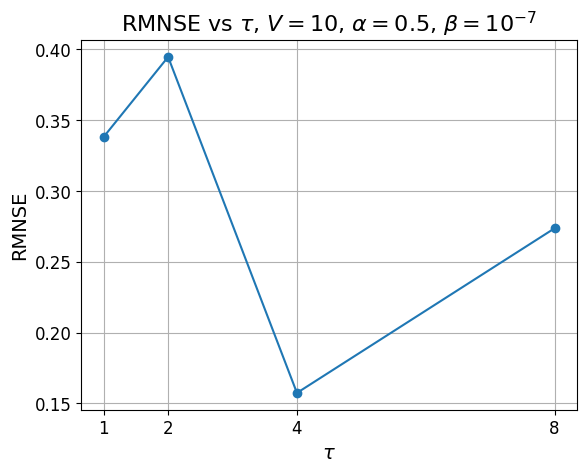}
        \caption{RMNSE error of modes vs timestep interval $\tau$}
        \label{fig:params-tau}
    \end{subfigure}
    \caption{HQRC- Comparison of RMNSE error. V=10, $\alpha = 0.5$, $\beta = 10^{-7}$.}
    \label{fig:params-hqrc}
\end{figure}

Figure \ref{fig:params-hqrc} compares RMNSE values for different parameters while keeping other parameters fixed. This justifies our choice of chosen hyperparameters.

\end{document}